# Misfire Detection in IC Engine using Kstar Algorithm

**Anish Bahri*, V.Sugumaran**, S. Babu Devasenapati*****

*SMBS, VIT University, Chennai Campus, Vandalur-kelambakam road, Chennai-600127, anish.bahri@gmail.com

**SMBS, VIT University, Chennai Campus, Vandalur-kelambakam road, Chennai-600127, v_sugu@yahoo.com

***Principal, Sri Subramanya College of Engineering and Technology, Palani, babudeva@yahoo.com

## ABSTRACT

Misfire in an IC Engine continues to be a problem leading to reduced fuel efficiency, increased power loss and emissions containing heavy concentration of hydrocarbons. Misfiring creates a unique vibration pattern attributed to a particular cylinder. Useful features can be extracted from these patterns and can be analyzed to detect misfire. Statistical features from these vibration signals were extracted. Out of these, useful features were identified using the J48 decision tree algorithm and selected features were used for classification using the Kstar algorithm. In this paper performance analysis of Kstar algorithm is presented.

## 1. INTRODUCTION

In case of misfire, uncombusted fuel accesses the exhaust system, where its combustion in the catalytic converter can cause a dangerous increase in temperature, damaging the catalytic converter due to thermal overload. It may also lead to increased emission of hydrocarbons. Thus it becomes necessary to detect misfire to prevent pollution and wastage of fuel. Methods using the torsional vibration signal of the crankshaft [1, 2] and the acceleration signal of the engine head [3-5] were developed to detect misfire. Klenk et al. used crankshaft speed [6] for misfire detection. Some works have already been carried out in this field using features such as cylinder deviation torque [8], instantaneous angular velocity [9] and instantaneous exhaust manifold pressure [10]. These methods can be costly and cumbersome. Machine learning approach has been used for fault diagnosis. Sugumaran et al. have used decision tree [11] and support vector machines [12] as classifiers. For fault diagnosis in bearings, Sugumaran et al. used support vector machines [13]. Sakthivel et al. used histogram features with decision tree [14]. Muralidharan et al. carried out fault diagnosis of monoblock centrifugal pump using

Naïve Bayes and Bayes Net classifier [15]. The use of engine vibration data requires minimum instrumentation. Machine learning when used with this data gives appreciable results and the system can be trained for changing conditions. Here the vibration signals from the engine head were recorded using an accelerometer and stored. Statistical features were derived from these signals and finally feature selection followed by feature classification was carried out. Here the feature selection was done using J48 decision tree algorithm and feature classification was carried out using Kstar algorithm.

## 2. EXPERIMENTAL SETUP

The experimental setup comprises mainly of the spark ignition IC engine test rig with provisions made in order to manually cause misfire in a particular cylinder and the data acquisition system. An accelerometer is attached on the engine which measures the vibration signals. The acquired signal is then passed through an ADC to collect the required data from which features are extracted. Fig.1 shows the basic flow of the steps involved in the whole process.

### 2.1 Engine Test Rig

Table 1: Specifications of the engine used

| FEATURES | SPECIFICATION |
|---|---|
| **Make** | Hindustan Motors |
| **Number of cylinders/stroke** | Four cylinders/four stroke |
| **Fuel** | Gasoline(Petrol) |
| **Rated power** | 7.35 kW |
| **Rated speed (alternator)** | 1500 rpm |
| **Engine stroke length** | 73.02 mm |
| **Engine bore diameter** | 88.9 mm |
| **Cooling** | Water cooled |

The engine test rig consists of a 10 hp four stroke vertical four cylinder engine. It has the provisions to simulate the misfire by cutting the electric supply to individual spark plugs. The engine accelerator is firmly attached at the desired position via screw and nut mechanism. The speed is measured and monitored by means of a tachometer. The specifications of the engine are provided in Table1.

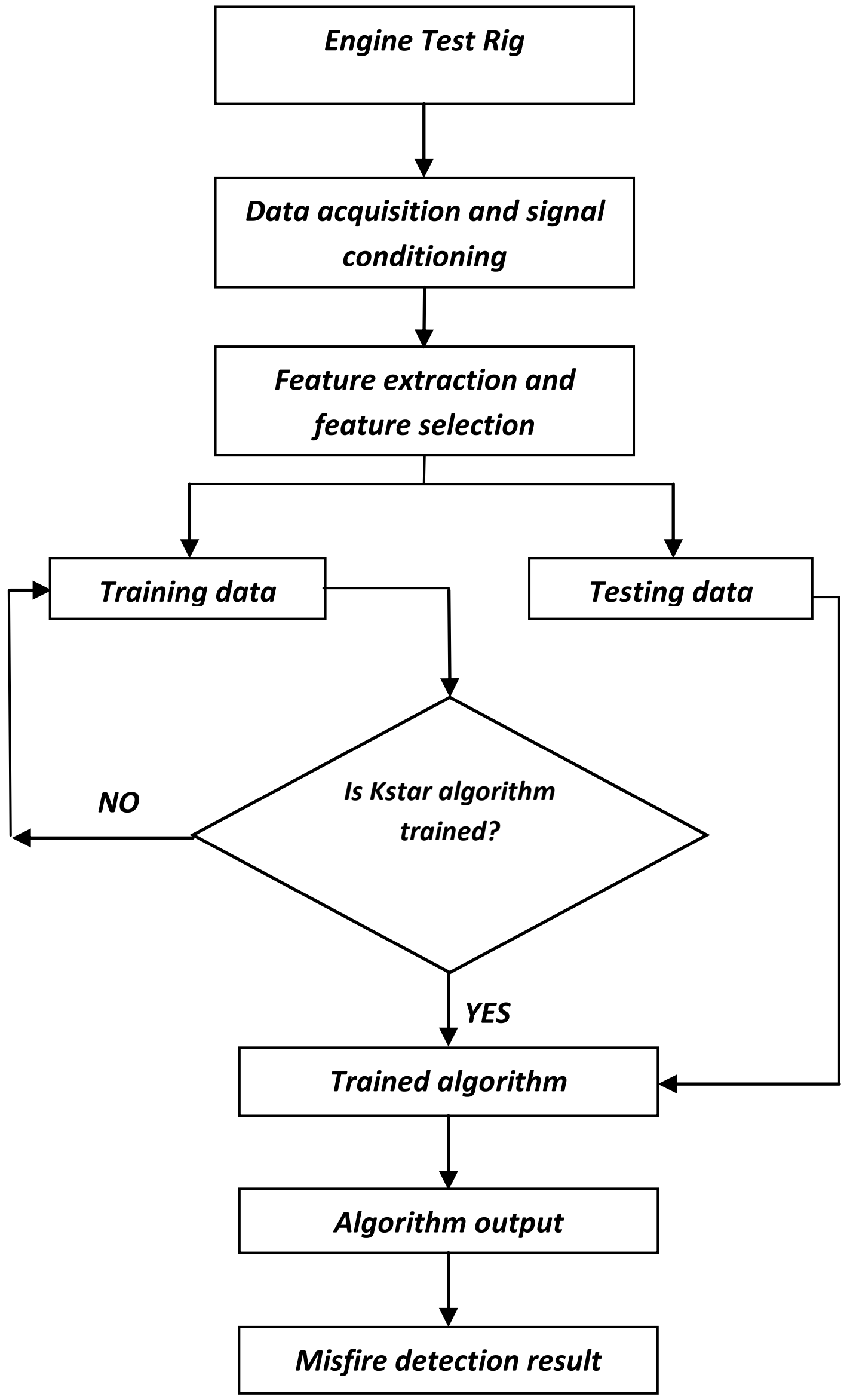

Fig. 1 Flowchart for Engine Misfire Detection

### 2.2 Data Acquisition System

Accelerometers were used to record the vibration signals. They have the ability to detect both small and large vibrations. The accelerometer is placed in such a way that it can measure vibration signals from all the four cylinders. The signal from the accelerometer is fed to a DACTRON FFT analyser which converts the analogue signal to digital signal. This data is transferred and stored in a computer for further processing. Fig 2 shows the experimental setup used.

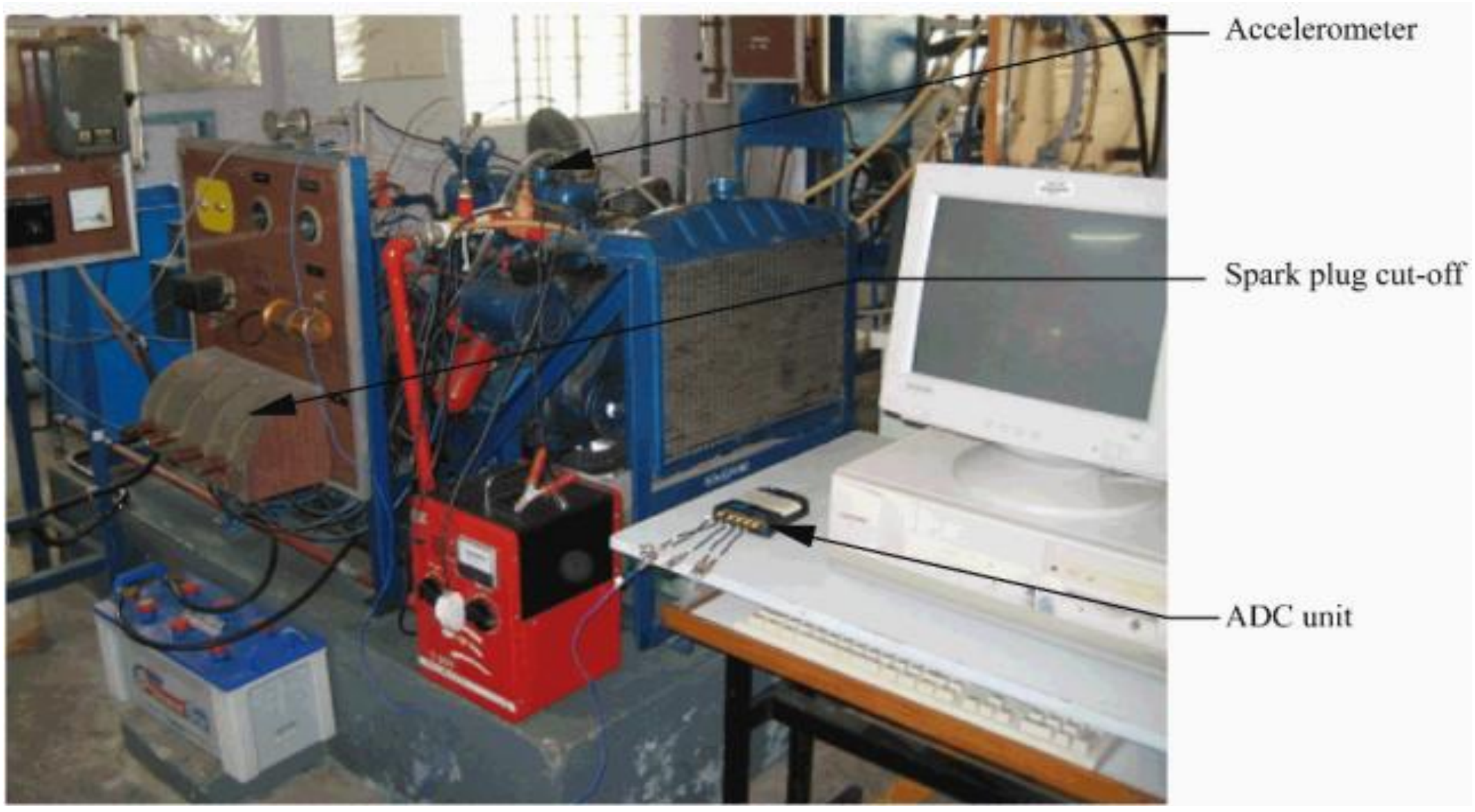

Fig 2 Experimental Setup

### 2.3 Experimental Procedure

Firstly the engine was started. This was done at no load by electrical cranking and warmed for 15 minutes. Now the FFT analyser was switched on and the data is taken only after the engine gets stabilised. All the data was collected for 1500 rpm, a sampling frequency of 24 kHz and a sampling length of 8192. For the present study five cases were considered i.e. normal condition, misfire in cylinder one, two, three and four. All the events were simulated at 1500 rpm. Fig. 3, 4, 5, 6 and 7 show vibration signals for no misfire, misfire in cylinder one, two, three and four respectively at 1500 rpm.

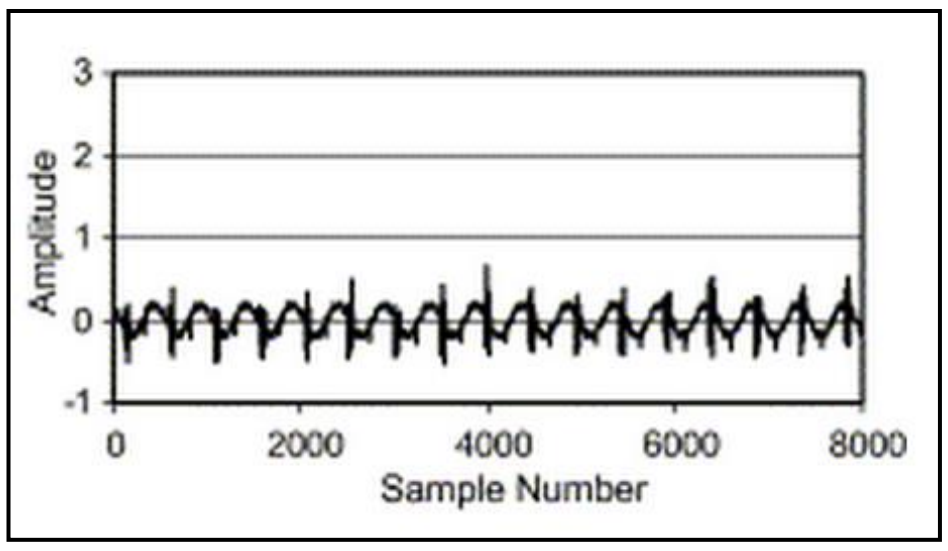

Fig. 3 Vibration signal for no misfire

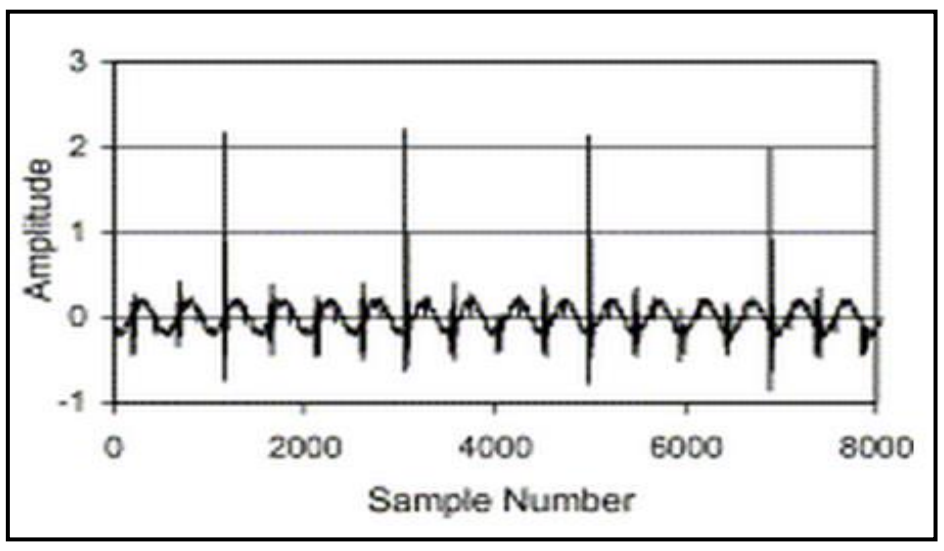

Fig. 4 Vibration signal for misfire in cylinder 1

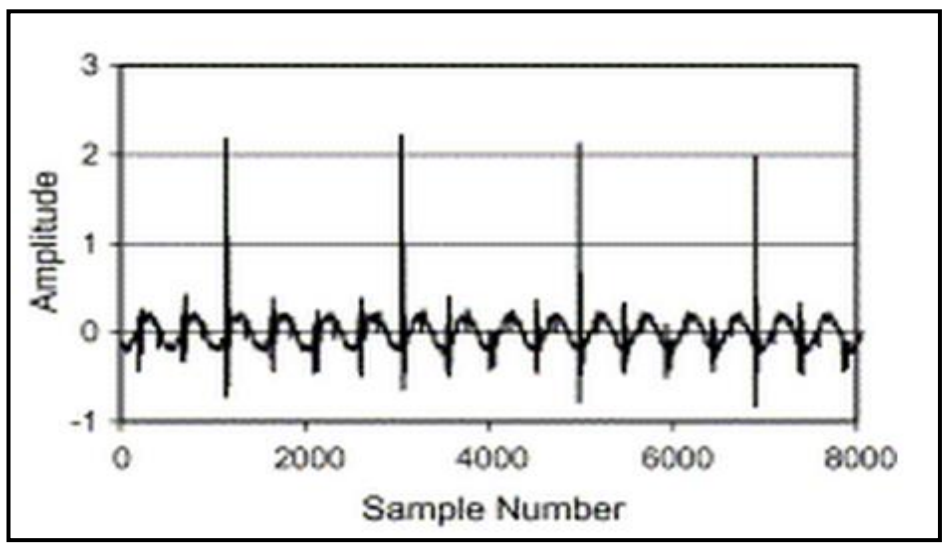

Fig. 5 Vibration signal for misfire in cylinder 2

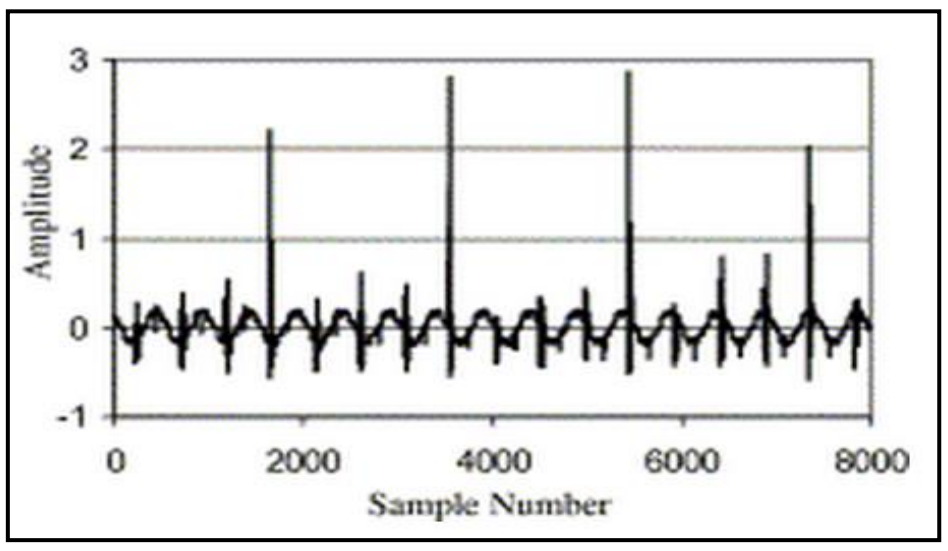

Fig. 6 Vibration signal for misfire in cylinder 3

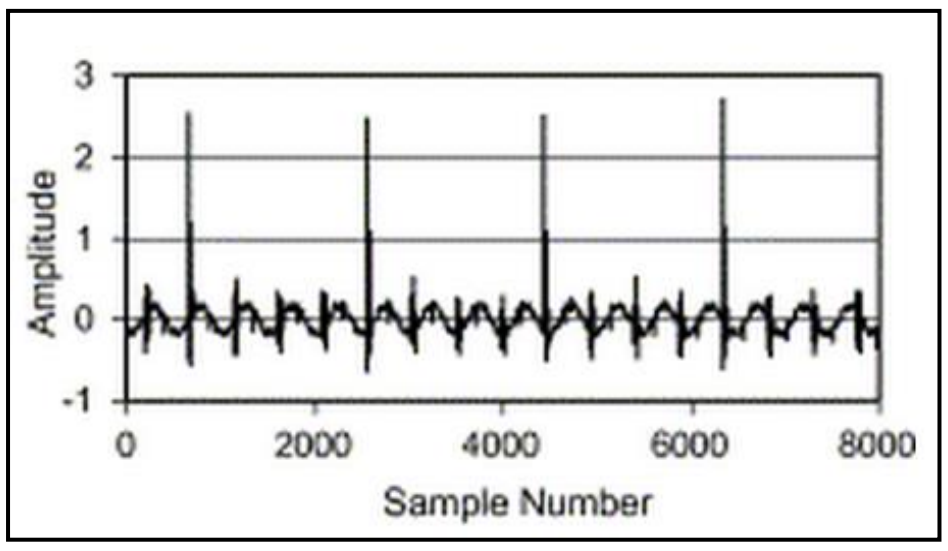

Fig. 7 Vibration signal for misfire in cylinder 4

## 3. FEATURE EXTRACTION AND FEATURE SELECTION

The vibration signals acquired cannot be used directly. Useful information needs to be extracted from them which will represent the signal. Here statistical features are extracted. Descriptive statistics for a particular signal gives a wide range of parameters namely mean, standard error, median, mode, standard deviation, sample variance, kurtosis, skewness, range, minimum, maximum, sum and count. The description of these parameters is given in Table 2.

Table 2: Description/Formulae of all statistical features

| Name of the feature | Description/Formulae |
|---|---|
| **Mean** | For a data set, mean is equal to the sum of all the data points divided by the number of data points. $\bar{x} = \frac{1}{n}\sum_{i=1}^{n} x_i$ |
| **Standard Error** | It refers to the error encountered when a statistic of sampling distribution varies from its value. $Standard\ errror = \frac{\sigma}{\sqrt{n}}$ |
| **Median** | Median is that value that separates the higher values from the lower values in a data sample. |
| **Mode** | It is that value that appears the most number of times in a data sample. |
| **Standard deviation** | It shows how much variation exists in the data from the average value. $Standard\ deviation\ (\sigma) = \sqrt{\frac{1}{N}\sum_{i-1}^{N}(x_i - \bar{x})^2}$ |
| **Sample Variance** | It is a measure of how far a set of data is spread out. $Sample\ Variance = \sigma^2$ |
| **Kurtosis** | **It is used to describe the distribution of given data around the mean.** $Kurtosis = \left\{\frac{n(n+1)}{(n-1)(n-2)(n-3)}\sum_{i=1}^{n}\left(\frac{x_i - \bar{x}}{s}\right)^4\right\} - \frac{3(n-1)^2}{(n-2)(n-3)}$ |
| **Skewness** | It describes asymmetry from the normal distribution in a set of statistical data. $Skewness = \frac{(n)}{(n-1)(n-2)}\sum_{i=1}^{n}\left(\frac{x_i - \bar{x}}{s}\right)^3$ |
| **Range** | In a set of data, range is the difference of highest and lowest values. |
| **Maximum Value** | It refers to the maximum value in a given data set. |
| **Minimum Value** | It refers to the minimum value in a given data set. |
| **Sum** | It refers to the sum of all feature values of a given sample. |
| **Count** | It refers to total number of data points. |

All the features extracted may not be useful for feature classification. Thus feature selection is carried out. Feature selection is the process of selecting the features from the above features which contribute highest to the classification accuracy. At the end of feature selection, only useful features will be left. This reduces the computational effort and thus reduces the cost and time. This is very important as the final aim is to convert this into an on-board diagnostic system.

For the present study, J48 decision tree algorithm was used for feature selection. Fig. 8 shows the decision tree obtained. It can be observed from the tree that root node is sample variance which is the feature contributing maximum to the classification accuracy closely followed by kurtosis and standard error. Coming down the tree, the following features continue, minimum, mean, standard deviation, skewness and range. Maximum, sum and count are not visible indicating that they do not contribute significantly to the classification accuracy.

After feature selection, the effect of number of features on classification accuracy was studied. Sample variance, standard error, kurtosis, minimum, mean, standard deviation, skewness and range were identified as the useful features and were thus used. Other non-relevant features were discarded.

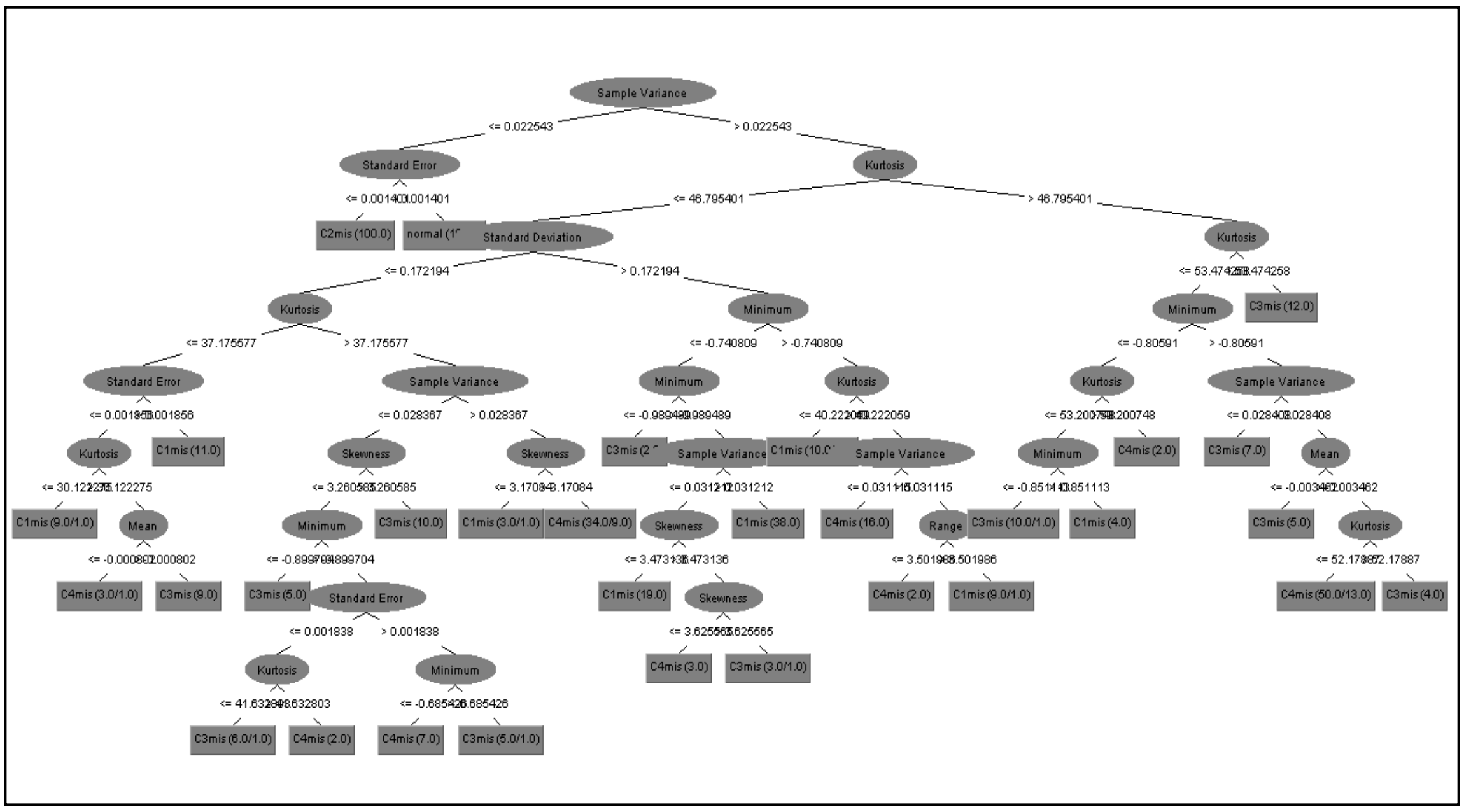

Fig. 8 Decision tree

## 4. CLASSIFIER

Here Kstar algorithm is used as the classifier. The Kstar algorithm uses entropic measure, based on probability of transforming instance into another by randomly choosing between all possible transformations. Using entropy as appraise of distance has numerous utility. Information theory helps in computing distance between instances. The complexity of a transformation of one instance into another is actually the distance between instances. This is achieved in two steps. First define a finite set of transformations they will map one instances to another. Then transform one instance (a) to another *(b)* with the help of "program" in a finite sequence of transformations starting at *a* and terminating at *b*.

4.1 Specification of Kstar

Given a set of infinite points and set of transformations predefined *T*. Let *t* be a value of set *T*. This *t* will map $t: I \rightarrow I$. To map instances with itself $\sigma$ is used in $T$ $(\sigma(a) = a)$. $\sigma$ terminates P, the set of all prefix codes from T*. Members of T* and of P uniquely define a transformation on I.

$\bar{t}(a) = t_n(t_{n-1}(\ldots t_1(a)\ldots))$ Where $t = t1 \ldots tn$

P is a probability function on T*. It satisfies the following properties:

$$0 \leq \frac{p(\bar{t}u)}{p(\bar{t})} \leq 1$$

$$\sum_u p(\bar{t}u) = p(\bar{t})$$

$$p(\Lambda) = 1$$

As a consequence, it satisfies the following:

$$\sum_{t \in P} p(\bar{t}) = 1$$

The probability function P* is defined as the probability of all paths from instance *a*, to instance *b*:

$$P*\left(\frac{b}{a}\right) = \sum_{t \in p: t(a)=b} P(t)$$

It is easily proven that P* satisfies the following properties.

$$\sum_b P*\left(\frac{b}{a}\right) = 1$$

$$0 \leq P*\left(\frac{b}{a}\right) \leq 1$$

The K* function is then defined as:

$K*(b/a) = -\log_2 P*\left(\frac{b}{a}\right)$. Please note that K* is not exactly a distance function.

## 5. RESULTS AND DISCUSSIONS

For misfire detection machine learning approach was used here. From a fairly large number of features, sample variance, standard error, kurtosis, minimum, mean, standard deviation, skewness and range were selected. The results are discussed below.

### 5.1 Effect of number of features on classification accuracy

All the features extracted may not be useful for feature classification. Thus the irrelevant features need to be filtered out. This was done using feature selection explained in earlier section. Table 3 shows the effect of number of features on classification accuracy.

Table 3: Classification accuracy of Kstar algorithm

| No. of features | Kstar algorithm accuracy (%) |
|---|---|
| **1** | 72.8 |
| **2** | 72.8 |
| **3** | 82.2 |
| **4** | 80.2 |
| **5** | 78.2 |
| **6** | 79.2 |
| **7** | 81.4 |
| **8** | **82.6** |
| **9** | 81.4 |
| **10** | 79 |
| **11** | 80.4 |
| **12** | 80.4 |

Thus it is observed that maximum classification accuracy is achieved using only 8 features out of 12 features. This reduces the computational effort and time.

### 5.2 Classification using Kstar algorithm

Table 4 Confusion matrix for Kstar algorithm

| *TESTING* | *C1mis* | *C2mis* | *C3mis* | *C4mis* | *Normal* |
|---|---|---|---|---|---|
| ***C1mis*** | *83* | *0* | *4* | *13* | *0* |
| ***C2mis*** | *0* | *100* | *0* | *0* | *0* |
| ***C3mis*** | *3* | *0* | *55* | *42* | *0* |
| ***C4mis*** | *3* | *0* | *22* | *75* | *0* |
| ***Normal*** | *0* | *0* | *0* | *0* | *100* |

Here results obtained from the Kstar algorithm will be discussed. Table 4 shows the confusion matrix obtained. In the confusion matrix, 'C1mis' stands for misfire in cylinder 1 and 'C2mis', 'C3mis','C4mis' stands for misfire in cylinders 2, 3 and 4 respectively. Normal means the engine is in normal conditions with no misfire. The first location in the first row shows the number of signals which are correctly classified as those from the first cylinder. The second position in the first row shows the number of signals which are wrongly classified as the signal from cylinder two. Similarly third and fourth positions show the number of signals wrongly classified as those from cylinder three and cylinder four. The first position in the second row shows the number of signals from cylinder two wrongly classified as those from cylinder one. The second position in second row shows the number of signals correctly classified as those from cylinder two. Thus the diagonal in the matrix shows the number of correctly classified points. It can be seen from the matrix that no faulty signal is classified as a normal signal which is very crucial. Thus Kstar algorithm can differentiate between good and faulty signals with 100% accuracy. Because of its ability to distinguish between the faulty and good signals with 100% accuracy no false alarm will be generated, thus saving time and resources which may be spent to check if the engine misfired. However there are some misclassifications among the faulty conditions as it is evident from Table 4. Kstar algorithm shows an overall classification accuracy of 82.6% which is appreciable because no time needs to be spent to pin point which cylinder misfired, and thus the problem can be directly rectified, thus saving money and resources.

## 6. Conclusion

Considering the emissions and fuel wastage caused, Kstar algorithm was proposed as a possible classifier for misfire detection using vibration signals from the engine block. Kstar performed satisfactorily with a classification accuracy of 82.6%. Also the ability to distinguish between faulty and good conditions with 100% accuracy makes it suitable for industry applications.